# Apricot variety classification using image processing and machine learning approaches

Seyed Vahid Mirnezami
Mechanical Engineering Department
Iowa State University
Ames, IA, USA
Vahid.gvg@gmail.com

Ali HamidiSepehr
Agricultural Engineering Department
North Carolina State University
Raleigh, NC, USA
ali.hamidisepehr@gmail.com

Mahdi Ghaebi
Agricultural Engineering Department
University of Tehran
mahdighaebi@yahoo.com

## ABSTRACT

Apricot which is a cultivated type of Zerdali (wild apricot) has an important place in human nutrition and its medical properties are essential for human health. The objective of this research was to obtain a model for apricot mass and separate apricot variety with image processing technology using external features of apricot fruit. In this study, five verities of apricot were used. In order to determine the size of the fruits, three mutually perpendicular axes were defined, length, width, and thickness. Measurements show that the effect of variety on all properties was statistically significant at the 1% probability level. Furthermore, there is no significant difference between the estimated dimensions by image processing approach and the actual dimensions. The developed system consists of a digital camera, a light diffusion chamber, a distance adjustment pedestal, and a personal computer. Images taken by the digital camera were stored as (RGB) for further analysis. The images were taken for a number of 49 samples of each cultivar in three directions. A linear equation is recommended to calculate the apricot mass based on the length and the width with $R^2 = 0.97$. In addition, ANFIS model with C-means was the best model for classifying the apricot varieties based on the physical features including length, width, thickness, mass, and projected area of three perpendicular surfaces. The accuracy of the model was 87.7.

## 1 INTRODUCTION

Apricot (Prunus Armeniaca L.) is a cultivated type of Zerdali (wild apricot) produced by inoculation. It has an important place in human nutrition and can be used as fresh, dried, or processed fruit [1]. Apricots are produced commercially in 63 countries on about 520455 hectares. Turkey, Uzbekistan, Italy, Algeria, Iran, Pakistan, Spain, France, Afghanistan, and Morocco are the top ten apricot producers in terms of annual production. In addition, its production exceeds four million ton (FAO, 2017).

Apricot is rich in minerals and vitamins, and it has been determined that the levels of vitamins and especially minerals in apricot fruits vary by different cultivation climate, variety, soil type, and ripening level at harvesting time [2], [3]. The final form of apricot variety matters commercially due to differences in customer tastes and physical properties. For example, dryness is one of the parameters that varies from one variety to another and is crucial for determining the commercial value of the product [4]. Hence, Grading and sorting different varieties play an important role in commercializing the final product. Moreover, obtaining information about the variety and physical properties can assist in sorting agricultural product [5].

Sorting the fruit varieties requires implementing the computational techniques including image processing and machine learning approaches known as remote sensing. Nowadays, computer sciences are getting more and more involved in agricultural and food science to make a decision based on estimated or actual parameters named as a feature [6]–[8]. Various artificial intelligence methods including machine vision and soft computing techniques have vast applications in fruit grading and to provide a higher quality product at the consumer end [9]–[11]. Image processing has been utilized as effective tools for measuring external features of fruits and plants such as color intensity, color homogeneity, bruises, size, shape, and stem identification [12], [13]. In addition, machine learning has been found increasingly useful in agricultural and food industry for applications in quality inspection, meeting quality standards and increasing market value [14]. Adaptive Neuro-Fuzzy Inference Systems (ANFIS) is one of the machine learning methods used for classification and regression in agricultural research [15], [16]. It is an adaptive model uses a hybrid learning with applying neural network together with fuzzy logic algorithm with its three different rule refinement methods including Grid Partitioning, Subtractive, and Fuzzy C-means clustering. [15], [17]. Generally, these methods have been implemented during recent decades in the agricultural applications to reduce labor intensive works and maximize the work efficiency [18].

Several types of research have been conducted using image processing to estimate different parameters in fruits including evaluating the maturity of banana [19], pawpaw [20], and orange [21]. In another research, watermelon quality was estimated based on its shape [22]. In addition, strawberry, apple, cranberry, and pomegranate were graded based on image processing [23]–[26]. In addition, an artificial neural network was utilized for different classification purposes. Recently, researchers have been widely implementing convolutional neural network known as deep learning to achieve the goals including fruit detection [27], [28] and classification [29], [30]. We did not use deep learning since it needs a large dataset even with using pre-trained network such as ImageNet [31]–[33]. In addition, there is a need to know to understand the features and interpret them for marketing issues. Recently However, no similar work has been done to classify the apricot varieties using either deep learning or conventional methods including image processing and machine learning methods.

Therefore, the objective of this research was to separate apricot fruit varieties with image processing and machine learning techniques specifically neural network and ANFIS. ANFIS was recommended by [16], [17] for agricultural product classification. In addition, it was recently used by some researchers to grade various fruits including pomegranate and apple [26], [34] based on physical properties. We extracted the external features of apricot fruits. The chemical, physical and mechanical properties of different apricot varieties have been determined by some researchers such as [2],

[35]–[37] as well. After that, a model was obtained for apricot mass based on the physical features obtained by image processing. Finally, the predicted mass along with the physical featurs were used to train different models for apricot variety classification. Considering classification apricot fruits and separating their varieties, they can be packaged according to regional characteristics for selling or types of customer's needs and increase producers share of the market. Separating apricot fruits by using image processing techniques can be reduced labor cost and time of the process and leads to increase quality and decrease waste of products.

## 2 Materials and Methods

### 2.1 Measuring actual size and mass

The apricot varieties used in this study were Ordubad, Shahroud, Maragheh, Oromieh, and Nasiri. They were supplied from Sahand Agricultural Research Center of Tabriz. These apricots were kept at the 5°C cooled box for experiments. The experiments were conducted at the physical properties laboratory, College of Abouraihan, University of Tehran, Pakdasht, Tehran, Iran. All dirt and damaged samples were removed to have a completely clean set. The moisture contents of fruits were determined by using an air oven method. Moreover, the oven temperature was set at 105±3°C and the samples weighted every 30 minutes until the weight difference in two consecutive weightings was less than 0.2% of the initial weight. Finally, 49 samples were selected for each apricot fruit cultivars which was a total of 245.

In order to determine the size of the fruits, three mutually perpendicular axes were shown in Figure 1. They are defined as length (L, the longest intercept along of pedicel), width (W, the longest intercept normal to L) and thickness (T, the longest intercept normal to L and W). The dimensions of each sample were measured along the axes by a micrometer with an accuracy of ±0.01 mm. The mass of each apricot fruit was measured by a digital balance with an accuracy of ±0.001 g. The dimensions and mass were measured for 245 total samples.

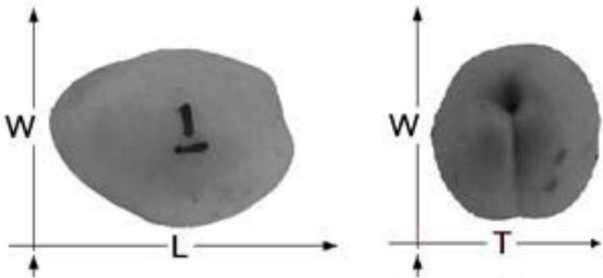

**Figure 1. Defined dimensions for apricot fruit. W: width, L: length, and T: thickness**

### 2.2 Experimental Setup and Image Acquisition

Figure 2 shows a developed pipeline for capturing the images of all samples. The image acquisition system consists of a digital camera (Canon Ixus 65, with 2816 by 2112-pixel resolutions), a light diffusion chamber, a distance adjustment pedestal, and a personal computer (PC). This system was used in order to lessen the potential errors in photography. supply controlled conditions consisted of:

**Environment lighting:** One fluorescent round tube was used to provide uniform lighting. The position of the tube was adjusted to provide uniform shadow and, well-illuminated images to avoid any shadowing effects on images.

**Distance:** A distance adjustment pedestal was used to obtain an equal distance between the camera and samples for all images. The equal distance was important to find a consistent calibration coefficient.

**Background:** A black cardboard was used as a background surface to facilitate and simplify the segmentation task. This is done because black and yellow known as fruits color have the highest contrast relative to each other.

**Camera vibration:** Vibration can cause impairment in the image acquisition, so the camera was fixed on the top of lightbox minimize variabilities during the photography process.

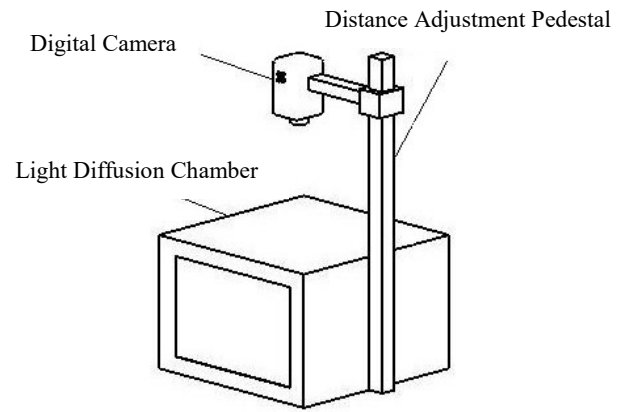

**Figure 2. Developed a photography system schematic**

### 2.3 Feature Extraction

As mentioned earlier, 49 apricots were selected from each variety. A distinct RGB picture was taken in three directions separately from each apricot which resulted in 735 images in total. MATLAB 2017b was used to extract the physical features of apricots. First, a binary thresholding method [38] was applied to convert the RGB images to binary. Figure 3 shows an apricot image before and after conversion.

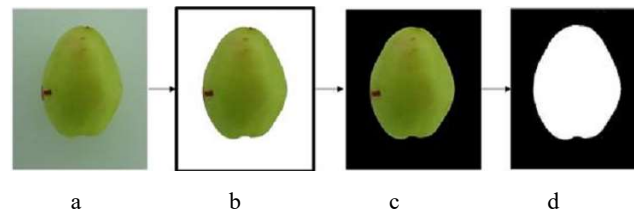

a     b     c     d

**Figure 3. converting RGB images to binary image. a) RGB image, b) foreground is removed, c) complement image, d) a binary image**

Then morphological algorithms were implemented to extract the physical features in terms of a number of pixels Initially, the projected area (PA) and dimensions of ''known'' materials were computed (Figure 4). To do so, the number of pixels in the foreground (fruit) can be scaled in order to convert (map) the total number of pixels into the actual dimensions. Accordingly, constant K (calibration coefficient), was obtained for each apricot fruit after rationing the actual and estimated area and dimensions of known samples. The outputs were compared with actual dimensions to validate the image processing approach. This validation method was also employed by previous researchers [39]–[41].

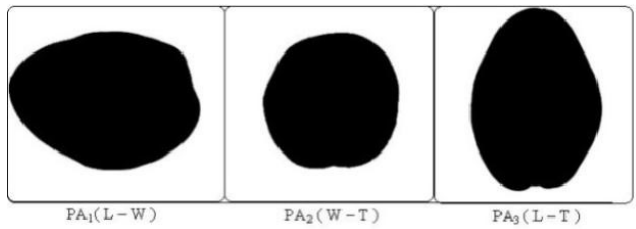

**Figure 4. Three mutually perpendicular axes used in image acquisition**

## 2.4 Mass Estimation

Fruit mass is one of the crucial features that can be useful for the classification of apricot varieties. Since the objective of this study is to develop a model for classification of the varieties, a method should be implemented to estimate the mass of the apricots. Therefore, the Linear Regression method was used to model the mass, using the six extracted physical features namely L, W, T, $PA_1$, $PA_2$, $PA_3$.

In order to find the best linear model (M) for estimating the mass of each sample, it was necessary to examine the different combinations of the features. To do this end, 63 models were created from a combination of the six features, and the accuracy was assessed for each model in Eq. (1).

$$M = W_0 + W_1F_1 + W_2F_2 + W_3F_3 + W_4F_4 + W_5F_5 + W_6F_6 \quad (1)$$

Where $F_1$ to $F_6$ are the features, and $W_1$ to $W_6$ are the weight of these features.

## 2.5 Modeling and Classification

There are various machine learning algorithms available for classification. In this paper, we used Multi-Layer Perceptron network (MLP), Radial Basis Functions network (RBF) and ANFIS suggested by [42], [43] and proved to be capable of producing accurate classification and prediction. The general parameters of the adopted one-layer MLP and RBF are mentioned in Table 1 and Table 2, respectively. ANFIS is an adaptive system which uses a hybrid learning algorithm to identify parameters of Sugeno-type fuzzy inference systems. Choosing an appropriate way for extracting the fuzzy rules is one of the critical steps in building an ANFIS network. In our study, we used three rule refinement methods including Subtractive Clustering, Grid Partitioning, and Fuzzy C-means proposed and utilized in [42], [43]. It is extremely important to tune the ANFIS training parameters to achieve the best classification results during designing of the ANFIS model. Our ANFIS parameters are given Table 3.

**Table 1. MLP Neural Network parameters**

| No. | Parameter/Method | Value/Description |
|---|---|---|
| 1 | Number of Hidden Layers | 1 |
| 2 | Number of Neurons in the hidden layers | 5 |
| 3 | Learning Algorithm | Levenberg-Marquart |
| 4 | Activation Function of Hidden Neurons | tanh (Sigmoid) |
| 5 | Activation Function of Output Neurons | Linear Transfer Function |
| 6 | Error Calculation Method | RMSE (Root-Mean-Square Error) |
| 7 | Maximum Learning epochs | 20 |
| 8 | Minimum Learning Error | $10^{-5}$ |
| 9 | Input Data Pre-Processing | 1. Remove Constant Input 2. Normalizing Inputs to the range of [0, 1] |

**Table 2. RBF Neural Network parameters**

| No. | Parameter/Method | Value/Description |
|---|---|---|
| 1 | Learning Algorithm | Least-Square |
|   | Standard Deviation or Distribution of data (σ) | 80 |
| 2 | Algorithm of Estimation of Hidden Neurons Number | Subtractive Clustering |
| 3 | Activation Function of Hidden Neurons | Radial Basis Function (Gaussian) |
| 4 | Activation Function of Output Neurons | Weighted Summation |
| 5 | Error Calculation Method | RMSE (Root-Mean-Square Error) |
| 6 | Maximum Learning epochs | 30 |
| 7 | Minimum Learning Error | 10-5 |
| 8 | Input Data Pre-Processing | 1. Removing Constant Input 2. Normalizing Inputs to the range of [0, 1] |

**Table 3. ANFIS parameters**

| No. | Parameter/Method | Value/Description |
|---|---|---|
| 1 | Learning Algorithm | Back-Propagation |
| 2 | Input Membership Function | Gaussian (gaussmf) |
| 3 | Output Membership Function | Linear Transfer Function |
| 4 | Error Calculation Method | RMSE (Root-Mean-Square Error) |
| 5 | Maximum Learning epochs | 20 |
| 6 | Minimum Learning Error | 10-5 |
| 7 | Input Data Pre-Processing | 1. Removing Constant Input 2. Normalizing Inputs to the range of [0, 1] |

## 3 Results and Discussion

### 3.1 Statistical Analysis of Varieties Dimension

Dimensional characteristics of fruits are shown in Table 4. Each value is the mean ± standard deviation of triplicate determinations. Means with different letters are significantly different ($p < 0.01$). The effect of variety for all dimensions was significant except the width of Maragheh and Oromieh. length, width, and thickness for all varieties are ranged from 34.87 to 46.64, 32.66 to 44.68, and 31.52 to 41.22mm, respectively. Based on the values, the greatest dimensional characteristics were found for Ordubad variety. Oromieh had the lowest length and width values, but the thickness of Maragheh was the lowest among the studied varieties. The mean values of length, width, and thickness for all samples, regardless of varieties were obtained 44.11, 40.04, and 37.96 mm, respectively. The order of these values confirmed by a researcher reported the means of length, width, and thickness for six cultivars of apricot 44.66, 41.31, and 38.95mm, respectively [37]. The axial dimensions are important for determining the aperture size of machines, particularly in materials separation and these dimensions may be useful in estimating the size of machine components.

**Table 4. Effect of variety on dimensional properties of apricot fruit**

| Variety | Ordubad | Shahrod | Maragheh | Oromieh | Nasiri |
|---|---|---|---|---|---|
| Length (mm) | 46.64 ±2.80[b] | 52.34 ±3.44[a] | 36.59 ±2.04[f] | 34.87± 1.93[g] | 45.62 ±3.07[c] |
| W(mm) | 44.68 ±3.11[c] | 38.44 ±2.71[e] | 33.22± 2.03[h] | 32.66± 1.93[h] | 42.83 ±3.39[d] |
| T(mm) | 41.22 ±2.53[d] | 38.39 ±2.74[e] | 31.52 ±1.77[i] | 32.51± 2.20[h] | 40.01 ±2.93[d] |

As shown in Figure 5, the correlation coefficients between the estimated values by image processing and actual values for length, width, and thickness of the apricots were 0.98, 0.96, and 0.98, respectively. The $R^2$ values can be interpreted as the proportion of the variance in the estimated values are attributed to the variance in the actual measurements. The higher values of the $R^2$ shows that the estimated results are closer to the actual results. The proposed image processing method yielded above 96% accuracy in estimating apricots fruits dimension.

According to the results of Table 5, the highest $PA_1$ and $PA_2$, values were for Ordubad, with a mean of 1878.12 and 1738.26 (mm$^2$), respectively. $PA_3$ of Shahrod (1856.98 mm$^2$) was significantly greater than the other ones. Oromieh had the lowest $PA_1$ and $PA_3$, with the values of 1062.29 and 1070.33 mm$^2$, respectively. The second and third projected areas ($PA_2$ and $PA_3$) for all of the varieties expect Shahrod was not significantly different at the 1% probability level. Hence, the effect of direction on Oromieh was not found significant.

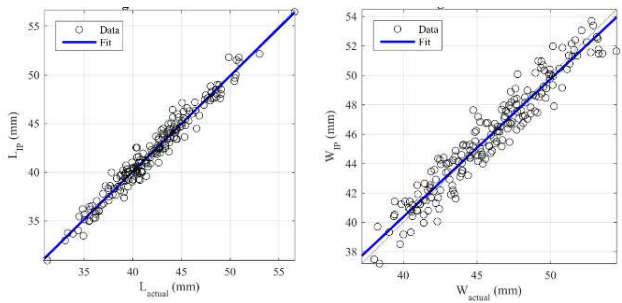

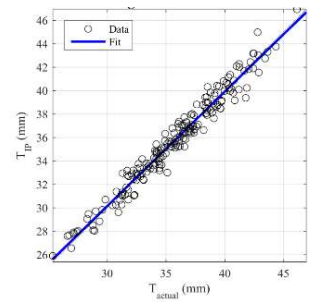

**Figure 5. Correlation coefficients between the estimated (IP) and actual (micrometer) dimensions of apricot fruits method**

**Table 5. Variations of the projected areas of apricot with variety and direction (Means with different letters are significantly different (p < 0.01))**

| Variety | Ordubad | Shahrod | Maragheh | Oromieh | Nasiri |
|---|---|---|---|---|---|
| $PA_1$ (mm2) | 1878.12 ±252.97[a] | 1860.30 ±254.04[a] | 1147.78 ±109.44[d] | 1062.29 ±103.95[de] | 1759.89 ±416.81[b] |
| $PA_2$ (mm2) | 1738.26 ±205.62[b] | 1459.64 ±227.65[c] | 1004.58 ±116.43[e] | 1015.78 ±116.02[e] | 1596.21 ±380.21[c] |
| $PA_3$ (mm2) | 1741.67 ±223.05[b] | 1856.98 ±251.28[a] | 1105.56 ±113.06[de] | 1070.33 ±115.80[de] | 1640.97 ±376.55[b] |

## 3.2 Statistical Analysis of Mass Estimation

The average and the standard deviation of actual measured mass for each category is mentioned in Table 6. Average of apricot fruits mass was between 21.39 and 48.17g. Among all of the apricot varieties, Ordubad has the highest fruit mass value and biggest size as discussed earlier. In addition, it is commonly accepted that size of the fruit affects consumer appeal and attractiveness. Hence, because of its bigger size, Ordubad variety is mostly preferred for fresh consumption.

**Table 6. Variety effect in apricot fruit mass (Means with different letters are significantly different (p < 0.01))**

| Variety | Ordubad | Shahrod | Maragheh | Oromieh | Nasiri |
|---|---|---|---|---|---|
| Average (g) | 47.81[a] | 38.36[b] | 22.51[c] | 21.39[c] | 44.69[b] |
| STD | 7.63 | 8.47 | 2.88 | 3.34 | 2.93 |

As discussed in the last section, a regression model was used to estimate the apricot mass. This model was selected among different combinations based on the lowest root mean square (RMS) and standard deviation error of the models. Therefore, the weights of the best model for estimating the mass of apricots is presented in Table 7. Also, evident is the fact that W1, W2, and W3 had the smallest values meaning that the L, W, and T features had the least effect on the modeling the mass.

**Table 7. weights of the best model for predicting the mass**

| Weight | $W_0$ | $W_1$ | $W_2$ | $W_3$ | $W_4$ | $W_5$ | $W_6$ |
|---|---|---|---|---|---|---|---|
| VALUE | −52.6035 | −0.0080 | 0.0006 | 0.0142 | 0.4752 | 0.4619 | 1.0814 |

The correlation, average error, standard deviation of error, and RMSE values for each variety are shown in Table 8. The correlation for each variety is high and more than 0.95 and confirms that the linear model is enough for predicting the mass of apricots. In addition, the average of error, standard deviation, and RMSE were 0.03, 1.85, and 1.85, respectively. Hence, there is no need to use non-linear or non-parametric models.

Table 8. Mass modeling results for each variety

|  | Ordubad | Shahrod | Maragheh | Oromieh | Nasiri | Mean | STD |
|---|---|---|---|---|---|---|---|
| Correlation ($R^2$) | 0.955 | 0.962 | 0.981 | 0.960 | 0.976 | 0.97 | 0.0112 |
| Average Error ($\bar{e}$) | 0.013 | 0.287 | 0.244 | -0.360 | -0.199 | -0.003 | 0.2791 |
| STD of Error | 1.600 | 2.624 | 1.748 | 1.868 | 1.401 | 1.85 | 0.4672 |
| RMSE ($\bar{e}$) | 1.585 | 2.612 | 1.747 | 1.883 | 1.401 | 1.845 | 0.4651 |

In addition, Figure 6-a shows the total correlation for a model when we test all 245 samples. In addition, Standard deviation of error histogram for the model is shown in the Figure 6-b and the average value is 1.8.

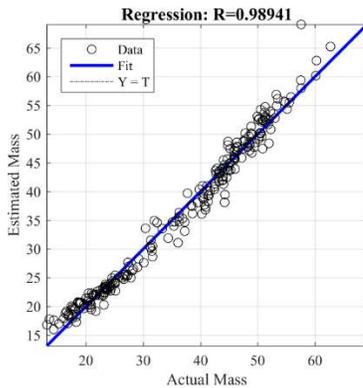
(a) The correlation coefficient between the estimated and actual mass values

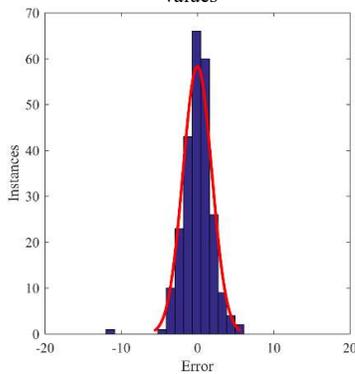
(b) Error Histogram
Figure 6. Mass modeling results

### 3.3 Classification Using ANN and ANFIS

As it is mentioned in the previous section, the estimated mass together with six extracted physical features were used to classify the five apricot varieties. The result of classifying using five classifiers including MLP and RBF networks as well as three ANFIS methods are mentioned in Table 9. It should be noted that in order to verify the performance of each classifier, 70% of the data were randomly selected for training, 15% for testing and 15% for verifying. and the figures mentioned in Table 9 are the averages of 10 times repeating each modeling.

As can be interpreted from the table, the overall accuracy of the MLP model was 83.6% which was higher than the RBF model. In addition, it can classify the apricots better for all varieties except the Ordubad. However, for each varieties the ANFIS models' performance were higher than both MLP and RBF. The accuracy of ANFIS models were 87.5%, 84.1%, and 87.7% for Grid Partitioning, Subtractive, and C-means, respectively. Based on the results, the ANFIS model with C-means clustering can conduct the classification better than the other modeling approaches.

Table 9. Results of classification with MLP, RBF, and ANFIS

|  | Ordubad (%) | Shahrod (%) | Maragheh (%) | Oromieh (%) | Nasiri (%) | Mean (%) |
|---|---|---|---|---|---|---|
| MLP | 77.1 | 85.5 | 82.7 | 84.0 | 82.6 | 83.6 |
| RBF | 77.3 | 81.8 | 79.8 | 79.6 | 79.8 | 80.6 |
| ANFIS (Grid Partitioning) | 85.1 | 88.2 | 86.9 | 85.3 | 86.6 | 87.5 |
| ANFIS (Subtractive Class) | 85.9 | 88.6 | 84.7 | 81.1 | 84.9 | 84.1 |
| ANFIS (C-means) | 80.6 | 87.2 | 85.1 | 84.5 | 85.0 | 87.7 |

## 4 Conclusion

The primary goal of this research was to an appropriate model able to classify five known varieties of apricot namely Ordubad, Shahroud, Maragheh, Oromieh, and Nasiri which are cultivated in subtropical countries like Iran. To do so, the linear regression model was chosen to estimate the mass of the apricot varieties using the 6 extracted features from the RGB images including 3 actual dimensions and the projected areas in xy, xz, and yz planes. The correlation for each variety was high and more than 0.95 and confirms that the linear model was enough for predicting the mass of apricots. In addition, the average of error, standard deviation, and RMSE were 0.03, 1.85, and 1.85, respectively.

For classification of different apricot varieties, five non-parametric modeling algorithms including MLP, RBF artificial neural networks and the ANFIS network with Grid-Partitioning, Subtractive, and C-means clustering approaches were used. Among these classifiers, ANFIS with C-means clustering and Grid Partitioning algorithms had performances with 87% accuracy in distinguishing the apricot types.